\documentclass[runningheads]{llncs}

\usepackage{eccv}

\usepackage{eccvabbrv}

\usepackage{graphicx}
\usepackage{subcaption}
\usepackage{booktabs}
\usepackage[percent]{overpic}

\usepackage[accsupp]{axessibility}  %

\usepackage{hyperref}
\newcommand{\ourmethod}{Generative Routing Pyramids\xspace}

\makeatletter
\DeclareRobustCommand\onedot{\futurelet\@let@token\@onedot}
\def\@onedot{\ifx\@let@token.\else.\null\fi\xspace}

\def\eg{\emph{e.g}\onedot}

 \def\vs{\emph{vs}\onedot}

\makeatother

\usepackage{orcidlink}

\usepackage{siunitx}

\begin{document}

\title{Unsupervised Learning of Cell Instances with Generative Routing Pyramids} 

\author{Ziwen Liu\inst{1}\orcidlink{0000-0001-7482-1299} \and
Martin Weigert\inst{1}\orcidlink{0000-0002-7780-9057}}

\authorrunning{Z.~Liu and M.~Weigert}
\institute{
Center for Scalable Data Analytics and Artificial Intelligence (ScaDS.AI), \\
Technische Universität Dresden, Germany \\
\email{\{ziwen.liu,martin.weigert\}@tu-dresden.de}
}

\maketitle

\begin{abstract}
    Identifying and representing object instances such as cells or nuclei
    is a common task in microscopy image analysis.
    Established machine learning workflows typically use supervised detection
    or segmentation followed by feature extraction or classification, 
    which requires manual annotations and treats instance segmentation
    and cell representation as separate stages.
    We describe a new unsupervised method for cell instance segmentation and
    phenotypic classification from unlabeled microscopy images.
    Our method is based on reconstructing each image using a coarse-to-fine routing pyramid
    that associates pixels with spatially sparse latent sources.
    The resulting pixel-to-latent associations yield instance masks,
    while the source latents encode cell morphology.
    We demonstrate competitive performance in instance segmentation
    across diverse cell morphologies and imaging modalities,
    as well as generative modeling of cellular phenotypes under perturbations.
    Source code and checkpoints are available at
    \url{https://github.com/weigertlab/routing-pyramids}.

\end{abstract}
\section{Introduction}
\label{sec:intro}

\begin{figure}[t]
  \centering
  \includegraphics[width=\textwidth]{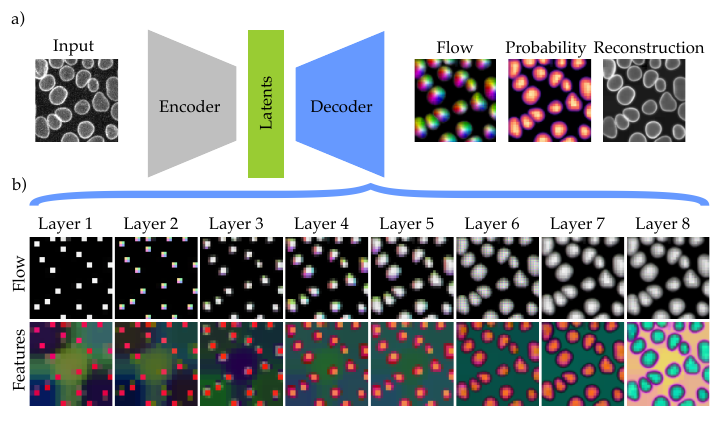}
  \caption{
    Overview of \ourmethod.
    a) From an input image,
    the encoder predicts a latent field,
    which is then used to generate pyramidal (coarse-to-fine)
    flows to reconstruct the image.
    b) Per-layer flow and routed feature maps.
    The flows are shown as HSV encoding,
    where hue is the flow direction,
    saturation is the flow magnitude,
    and value is the routed probability mass.
    The features are shown as RGB values representing
    the first three principle components.
  }
  \label{fig:overview}
\end{figure}
 
Identifying individual objects, such as cells or nuclei, in microscopy images
is a fundamental task in quantitative bioimage analysis~\cite{meijering2012cell}
with applications including cell counting, morphological profiling,
and tracking~\cite{boutros2015,amat2014,ulman2017}.
Currently, supervised deep learning-based approaches
are the most successful cell segmentation methods
\cite{ronnebergerUNet2015,schmidtCell2018,stringerCellpose2021,pachitariuCellposeSAM2025}.
Training these models, however,
requires laborious manual annotations of large sets of training images,
making dataset creation a recurring cost when specimens,
imaging modalities, or acquisition conditions change.

Although object appearances can vary substantially across imaging modalities and sample types,
they often exhibit a more stereotypic visual appearance within a single experimental setting,
particularly in cell culture experiments such as those considered here.
Such settings provide many unlabeled examples from which a model could learn shared structure
while retaining object-specific variation.
Yet a shared appearance model alone does not establish instance identity,
as adjacent cells can look similar but must remain associated with distinct
spatial sources.
Label-free cell instance learning therefore requires a shared appearance model
together with a spatial assignment that keeps individual objects separate.
Object-centric learning provides a framework for this idea:
reconstruction-based approaches explain an image as a collection of objects
and background and learn an object-level latent representation for each component
\cite{burgessMONet2019,greffMultiObject2020,locatelloObjectCentric2020,linSPACE2020}.
Existing formulations commonly separate components through global slots,
iterative inference, or independently rendered object glimpses,
but scale poorly to microscopy images with many small objects
(\cref{sec:related-object}).

In this work, we introduce \emph{\ourmethod},
an object-centric generative model trained directly on unlabeled microscopy images.
The model predicts where objects are present and runs a single decoder pass across
all candidate objects to reconstruct the image from their latent representations.
Because foreground presence is penalized, the model activates a source only
when the shared decoder can use it to reconstruct a recurring object appearance.
To reconstruct extended objects while preserving their spatial identity,
the decoder expands each source through a coarse-to-fine pyramid.
At each level, every finer-scale location softly chooses among nearby parent locations,
so the resulting chain of local choices traces image pixels back to
the sources that explain them.
These pixel-to-source ancestries yield instance masks,
while the corresponding source latents provide object representations
for single-cell phenotype analysis.
\section{Related Work}
\label{sec:related}

\subsection{Cell Segmentation and Representation Learning}
    \label{sec:related-cell}
    Supervised cell instance segmentation models often predict
    geometric targets derived from instance masks.
    StarDist~\cite{schmidtCell2018} regresses radial distances along a fixed set of
    directions to form star-convex polygon proposals,
    followed by non-maximum suppression.
    Cellpose~\cite{stringerCellpose2021} instead predicts dense pixel-space flows toward object centers
    and groups pixels by following these flows over multiple steps.
    Our method instead trains without flow targets or instance annotations.

    Unsupervised methods replace mask-derived targets with surrogate objectives.
    Cellulus~\cite{wolfUnsupervised2023} predicts relative offsets between image
    patches to learn dense spatial embeddings,
    estimates foreground by comparing embeddings of corrupted input images,
    and obtains instances through mean-shift clustering and morphological filtering.
    CellSeg3D~\cite{achardCellSeg3D2025} adapts W-Net~\cite{xiaWNet2017}
    to learn semantic foreground without labels
    and separates instances with Voronoi-Otsu post-processing.
    Both Cellulus and CellSeg3D separate instances via non-learned post-processing.
    Motion-based pseudo-labels provide another alternative~\cite{robitailleSelfsupervised2022},
    but require time-lapse data with smooth foreground motion and static background.
    Our method works with static images,
    and directly learns instance assignment as part of the image generation process.

    Microscopy representation learning either produces dense pixel or patch features
    \cite{gallusserSelfsupervised2023,krausMasked2024},
    which still require instance extraction for single-cell analysis,
    or learns from single-cell crops~\cite{ulicnaLearning2023,hirata-miyasakiDynaCLR2025},
    which assume prior detections or masks.
    In both cases, segmentation and representation learning are treated as distinct stages,
    requiring separate data curation and training efforts.
    Our method instead jointly learns instance segmentation and object representation
    from the same datasets without labels.

\begin{table}
  \centering
  \small
  \caption{Inference scaling of related methods. $N$ is the number of pixels,
  with feature-token count linear in $N$; $K$ is the object, slot, or seed
  component count; $T$ is the number of iterative inference steps; $R$ is the
  number of noisy inference passes; and $I$ is the number of mean-shift
  iterations. Fixed network-depth and local-neighborhood factors are omitted.
  \textsuperscript{$\dagger$}Cellulus does not reconstruct the image,
  its instance-segmentation cost is shown instead.}
  \begin{tabular}{@{}p{0.25\linewidth}p{0.25\linewidth}p{0.44\linewidth}@{}}
    \toprule
    Detection & Reconstruction & Method \\
    \midrule
    $O(KN)$, sequential
      & $O(KN)$
      & MONet~\cite{burgessMONet2019} \\
    $O(TKN)$
      & $O(TKN)$
      & IODINE~\cite{greffMultiObject2020} \\
    $O(TKN)$
      & $O(KN)$
      & Slot Attention~\cite{locatelloObjectCentric2020} \\
    $O(N^2+TKN)$
      & $O(N^2+KN)$
      & DINOSAUR~\cite{seitzerBridging2023} \\
    $O(N)$
      & Up to $O(N^2)$
      & SPACE~\cite{linSPACE2020} \\
    $O(RN)$
      & $O(IN\log N+KN)$
      & Cellulus\textsuperscript{$\dagger$}~\cite{wolfUnsupervised2023} \\
    $O(N)$
      & $O(N)$
      & CellSeg3D~\cite{achardCellSeg3D2025}
      (W-Net~\cite{xiaWNet2017}) \\
    $O(N)$
      & $O(N)$
      & \ourmethod (ours) \\
    \bottomrule
  \end{tabular}
  \label{tab:object-centric-scaling}
\end{table}
 \subsection{Object-Centric Generative Learning}
    \label{sec:related-object}
    \emph{Object-centric} generative models learn component representations through image
    reconstruction, but differ in how they infer and render components~\cite{burgessMONet2019,
    greffMultiObject2020,locatelloObjectCentric2020,linSPACE2020,seitzerBridging2023}. Their
    computational scaling matters for microscopy images containing many small
    instances (\cref{tab:object-centric-scaling}).
    Global-component methods couple full-image cost (with $N$ pixels) to the component count $K$,
    and iterative variants add the factor $T$.
    For example, SPACE~\cite{linSPACE2020} performs parallel candidate inference spatially,
    but renders every proposal in full image space.
    Our method predicts all candidate sources in one $O(N)$ forward pass,
    and reconstructs a shared spatial feature field instead of rendering each candidate separately.
    Reading out the $K$ inferred instance masks costs $O(KN)$.

\subsection{Hierarchical Flow Fields}
    \label{sec:related-hierarchy}
    Coarse-to-fine pyramids are established tools for approximating displacement fields
    \cite{burtLaplacian1983,bergenHierarchical1992} efficiently,
    and learned pyramids have been used to predict optical flow between video frames
    \cite{ranjanOptical2016}.
    In our setting, a coarse-to-fine hierarchy defines the spatial information flow
    of a single-image generative model, which induces instance segmentation.
    Intuitively, this spatial flow field
    is similar to the object center-to-boundary flows found in StarDist and Cellpose,
    but is implicitly defined by the image generation process.
    Using a hierarchical approximation avoids non-differentiable assignment like StarDist,
    or running many Cellpose-style flow simulation steps during training,
    allowing for end-to-end unsupervised learning.
\section{Method}
\label{sec:method}

\subsection{Overview}
\label{sec:method-overview}

Our method learns object instances from unlabeled microscopy images.
As shown in \cref{fig:overview},
an encoder infers foreground and background latents given an input image,
while a sparsity-regularized presence map marks object candidates.
A coarse-to-fine pyramidal decoder reconstructs
the image using local routing weights.
The same weights associate pixels with latent sources,
yielding instance masks and one latent representation per inferred object.
For timelapse data, the model processes each frame independently.

\subsection{Spatial Latents and Learned Object Seeds}
\label{sec:method-latents}

Given an input image $x\in\mathbb{R}^{H\times W\times C}$,
the encoder $E_\phi$ produces features
$f=E_\phi(x) \in \mathbb{R}^{h \times w \times d_E}$ on a grid
$\Omega_0$ with spatial stride $\delta$.
The stride is a model hyperparameter,
with $h=H/\delta$ and $w=W/\delta$ for divisible input sizes.
We use $\delta=8$ in all experiments.

Pointwise heads predict a diagonal Gaussian posterior
for each of the two latent grids,
\begin{equation}
    q^r_\phi(z^r\mid x)
    =\prod_{u\in\Omega_0}
    \mathcal{N}\!\left(z^r_u;\mu^r_u,
    \operatorname{diag}((\sigma^r_u)^2)\right),
    \qquad r\in\{\mathrm{fg},\mathrm{bg}\}
\end{equation}
Here $z^r_u\in\mathbb{R}^d$, where $d$ is the shared latent dimension.
The foreground and background candidates are sampled independently given
$x$, and both use standard-normal priors. Before the background head,
encoder features are average-pooled with stride $\delta_{\mathrm{bg}}$ and
bilinearly restored to $\Omega_0$. This stride is a separate model
hyperparameter; we use $\delta_{\mathrm{bg}}=8$ in all experiments.
This operation restricts the image-dependent background posterior parameters
to vary smoothly across space.

A pointwise prediction from the foreground candidate defines a deterministic
soft presence gate
\begin{equation}
    s_u=\operatorname{sigmoid}(w_s^\mathsf{T}z^{\mathrm{fg}}_u+b_s),
    \qquad u\in\Omega_0
\end{equation}
The decoder starts from the gated blend
\begin{equation}
    h^0_u=s_u z^{\mathrm{fg}}_u+(1-s_u)z^{\mathrm{bg}}_u
\end{equation}
Values of $s_u$ near one expose the foreground candidate, whereas values
near zero select the spatially smoothed background candidate. The posterior
candidates are sampled independently, but this deterministic gate couples
them before decoding.
During training, the Gaussian fields are sampled via reparameterization
\cite{kingmaAutoEncoding2014}, and inference uses their posterior means.
Reconstruction pressure determines where foreground capacity is needed,
and sparse active regions of $s$ are interpreted as learned object seeds.

\subsection{Routing Pyramid Decoder}
\label{sec:method-decoder}

Let $\Omega_\ell$ denote the destination grid at decoder layer $\ell$.
Each site $j\in\Omega_\ell$ selects among a valid $3\times3$ neighborhood
$\mathcal{N}_\ell(j)$ around its anchor on the preceding parent grid, which
is either at the same resolution or coarser by a factor of two. A
convolutional residual predictor produces logits $a^\ell_{ji}$ from the
presence-gated decoder features, and a masked softmax gives
\begin{equation}
    T^\ell_{ji}
    =\frac{\exp(a^\ell_{ji})}
    {\sum_{i'\in\mathcal{N}_\ell(j)}\exp(a^\ell_{ji'})},
    \qquad i\in\mathcal{N}_\ell(j),
    \qquad \sum_i T^\ell_{ji}=1
\end{equation}
Invalid boundary edges have zero probability.
Thus $T^\ell_{ji}$ is the conditional probability
that destination $j$ selects site $i$ as its parent.
Under a uniform measure over $\Omega_\ell$,
$T^\ell$ can therefore be viewed as a locally supported,
one-sided transport coupling.

The layer-indexed pyramid sites and their local parent links form
a directed acyclic graph (DAG) $G$:
an edge connects $(\ell,j)$ to $(\ell-1,i)$ whenever $i\in\mathcal{N}_\ell(j)$.
Every edge therefore points from output pixels
toward the latent seed grid and decreases the layer index.
Conditioned on the inferred image representation,
the locally normalized routing distributions define
a finite, layer-inhomogeneous Markov chain over $G$.
If assignment mass reaches a site $j$,
the weights $T^\ell_{ji}$ distribute all of it among valid parents;
flows can split at fine sites and merge at parent sites.
Composing the chain's layer-wise transition kernels gives
\begin{equation}
    A_{nu}
    =\sum_{\gamma\in\Gamma(n,u)}
    \prod_{\ell}T^\ell_{\gamma_\ell\gamma_{\ell-1}}
\end{equation}
where $n$ indexes output pixels and $\Gamma(n,u)$ is the set of valid paths
from pixel $n$ to coarse seed site $u$. Hence $A_{nu}$ is the probability
that unit assignment mass originating at $n$ reaches $u$, and
$\sum_u A_{nu}=1$. Thus
$A\in[0,1]^{HW\times hw}$ is the pixel-to-seed association matrix induced by
the pyramid DAG, with output pixels as rows and latent seed sites as columns.
Each row is the full transition distribution from one pixel to the seed
grid. Equivalently, it describes the unit flow induced from each pixel
through the learned probabilistic flow network.
This Markov-chain interpretation applies to the ancestry process after the
image-dependent routing kernels have been predicted;
feature propagation additionally includes the neural updates described below.
The propagated foreground probability is
\begin{equation}
    p^{\mathrm{fg}}_n=\sum_{u\in\Omega_0}A_{nu}s_u
\end{equation}

During generation, the same routing weights pull seed presence and decoder
values from each destination's parents,
\begin{equation}
    s^\ell_j=\sum_i T^\ell_{ji}s^{\ell-1}_i,
    \qquad
    r^\ell_j=\sum_i T^\ell_{ji}V_\ell(h^{\ell-1}_i)
\end{equation}
The transported update $r^\ell_j$ is normalized, activated, gated by
$s^\ell_j$, and added to a nearest-neighbor residual path. A pointwise
feed-forward residual block completes the layer. Layers either remain at
the current resolution or upsample by two; a final pointwise network maps
the full-resolution features to the reconstructed image $\hat{x}$. Thus the
coarse-to-fine generative computation and the fine-to-coarse ancestry
interpretation use the same learned graph.

Let $c_u$ be the image-space center coordinate of seed site $u$ and $c_n$
the coordinate of pixel $n$. The expected pixel-to-seed flow is
\begin{equation}
    v_n=\sum_{u\in\Omega_0}A_{nu}c_u-c_n
\end{equation}
This expected displacement gives a Cellpose-like flow readout from the
learned association matrix~\cite{stringerCellpose2021}.
Hard local support at every layer bounds individual transport steps, while
their composition permits larger object extents through the pyramid.

\subsection{Unsupervised Learning Objective}
\label{sec:method-objective}

Let $z=(z^{\mathrm{fg}},z^{\mathrm{bg}})$ be the sampled latent fields
and $\hat{x}_\theta(z)$ be the corresponding decoder output.
The expected reconstruction term is
\begin{equation}
    \mathcal{L}_{\mathrm{rec}}(x)
    =\mathbb{E}_{q_\phi(z\mid x)}
    \left[\frac{1}{HWC}
    \lVert\hat{x}_\theta(z)-x\rVert_1\right]
\end{equation}
Separate mean KL terms regularize the foreground and background posteriors
toward their standard-normal priors:
\begin{equation}
    \mathcal{L}^{r}_{\mathrm{KL}}
    =\frac{1}{|\Omega_0|d}\sum_{u,k}
    \frac{1}{2}\left((\mu^r_{uk})^2+(\sigma^r_{uk})^2
    -1-\log (\sigma^r_{uk})^2\right),
    \qquad r\in\{\mathrm{fg},\mathrm{bg}\}
\end{equation}

A smooth concave penalty encourages foreground capacity to concentrate into
fewer seed regions,
\begin{equation}
    \mathcal{L}_{\mathrm{sparsity}}
    =\frac{1}{|\Omega_0|}\sum_{u\in\Omega_0}
    \left((s_u+\epsilon)^\alpha-\epsilon^\alpha\right),
    \qquad \alpha\in(0,1)
\end{equation}
The exponent $\alpha$ is a hyperparameter controlling the concavity
of the penalty; we use $\alpha=0.5$ in all experiments.
Here $\epsilon$ is a small offset that keeps the derivative finite near $s_u=0$.

With $c^{\ell-1}_i$ and $c^\ell_j$ denoting parent- and destination-site
centers in image pixels, each local edge has quadratic geometric cost
$\lVert c^{\ell-1}_i-c^\ell_j\rVert_2^2$. We regularize its expectation
under the learned routing distributions,
\begin{equation}
    \mathcal{L}_{\mathrm{flow}}
    =\frac{1}{\sum_{\ell=1}^{L}|\Omega_\ell|}
    \sum_{\ell=1}^{L}\sum_{j\in\Omega_\ell}
    \sum_{i\in\mathcal{N}_\ell(j)}
    T^\ell_{ji}\lVert c^{\ell-1}_i-c^\ell_j\rVert_2^2
\end{equation}
Thus $\mathcal{L}_{\mathrm{flow}}$ averages the expected squared local step
length across layers and destination sites. It is generally different from
the squared norm of the composed expected flow $v_n$.

The complete per-image objective is
\begin{equation}
    \mathcal{L}
    =\lambda_{\mathrm{rec}}\mathcal{L}_{\mathrm{rec}}
    +\lambda_{\mathrm{fg}}\mathcal{L}^{\mathrm{fg}}_{\mathrm{KL}}
    +\lambda_{\mathrm{bg}}\mathcal{L}^{\mathrm{bg}}_{\mathrm{KL}}
    +\lambda_{\mathrm{flow}}\mathcal{L}_{\mathrm{flow}}
    +\lambda_{\mathrm{sparsity}}\mathcal{L}_{\mathrm{sparsity}}
\end{equation}
We use $\lambda_{\mathrm{rec}}=1.0$,
$\lambda_{\mathrm{fg}}=0.01$,
$\lambda_{\mathrm{bg}}=0.05$,
$\lambda_{\mathrm{flow}}=0.005$
for all experiments,
and choose $\lambda_{\mathrm{sparsity}}\in\{0.2, 0.5\}$
for different datasets (\cref{sec:experiments}).
We use linear warmup for the weight of regularization loss terms
over the first 10 epochs.

\subsection{Instance Segmentation and Object Representations}
\label{sec:method-inference}

At inference, the model replaces the sampled latent fields with their
posterior means. It thresholds the coarse presence map at $\tau_s$ and groups
neighboring active source sites using 8-connectivity:
\begin{equation}
    \{C_k\}=\operatorname{CC}_8(\{u:s_u\geq\tau_s\})
\end{equation}
Each component serves as an object seed. We use $\tau_s=0.5$ in all
experiments.

The routing ancestry expands these coarse seeds into image-space instances.
The presence-weighted mass of component $C_k$ at pixel $n$ is
\begin{equation}
    m_k(n)=\sum_{u\in C_k}A_{nu}s_u.
\end{equation}
A pixel is assigned to the component with the largest mass if that mass
reaches the threshold $\tau_m$:
\begin{equation}
    \hat{y}_n=
    \begin{cases}
        \operatorname*{arg\,max}_k m_k(n),
            & \max_k m_k(n)\geq\tau_m,\\
        0, & \text{otherwise}
    \end{cases}
\end{equation}
We use $\tau_m=0.1$ in all experiments.
Instances are filtered by minimum area on the model-resolution grid.
We use a minimum area of 100 pixels for all experiments.
The masks are then rescaled by nearest-neighbor (for upsampling)
or mode pooling (for downsampling) to the raw image resolution
if the model and raw image resolutions differ.

The same seed components define object-level representations
used for phenotype analysis.
For each component, its embedding $e_k$ is
the presence-weighted average of the foreground posterior means:
\begin{equation}
    e_k=\frac{\sum_{u\in C_k}s_u\mu^{\mathrm{fg}}_u}
    {\sum_{u\in C_k}s_u}
\end{equation}
\section{Experiments}
\label{sec:experiments}

\subsection{Unsupervised Cell Instance Segmentation}
\label{sec:segmentation-experiments}

We test whether the generative objective and inductive biases of our method
are sufficient for cell detection and instance segmentation without annotations.

\begin{table}[t]
    \caption{
        Biological and imaging properties of the instance-segmentation datasets.
    }
    \label{tab:segmentation-datasets}
    \centering
    \small
    \begin{tabular}{@{}p{0.15\linewidth}p{0.35\linewidth}p{0.25\linewidth}p{0.2\linewidth}@{}}
        \toprule
        \raggedright Dataset
            & \raggedright Cell type
            & \raggedright Visible structure
            & \raggedright Modality \tabularnewline
        \midrule
        \raggedright Allen
            & \raggedright Human induced pluripotent stem cells (hiPSCs)
            & \raggedright Nuclear envelope (Lamin B1-mEGFP)
            & \raggedright Fluorescence \tabularnewline
        \raggedright Fluo-HeLa
            & \raggedright HeLa cells
            & \raggedright Nuclei (H2B-GFP)
            & \raggedright Fluorescence \tabularnewline
        \raggedright PhC-PSC
            & \raggedright Rat pancreatic stem cells
            & \raggedright Cell body
            & \raggedright Phase contrast \tabularnewline
        \bottomrule
    \end{tabular}
\end{table} %
\begin{table}[tb]
  \caption{
    Instance segmentation on Allen, Fluo-HeLa, and PhC-PSC datasets.
    We compare the unsupervised methods Routing Pyramids, Cellulus, and Otsu
    with the supervised Cellpose-SAM reference.
    We report instance-level $F_1$ scores at IoU thresholds of 0.5, 0.7, and 0.9,
    and panoptic quality (PQ).
    Higher values are better.
    Bold and underlined values indicate the best and second-best result
    for each dataset and metric, respectively.}
  \label{tab:seg_results}
  \centering

  \begin{tabular}{l@{\hspace{5mm}}llrrrr}
    \toprule
    Dataset & Method & Type & $F_1^{[0.5]}$ & $F_1^{[0.7]}$ & $F_1^{[0.9]}$ & PQ \\
    \midrule
    Allen
    & Cellpose-SAM & {\scriptsize supervised} & \textbf{0.977} & \textbf{0.964} & 0.256 & \underline{0.859} \\
    & Otsu (size 42) & {\scriptsize non-parametric} & 0.906 & 0.861 & 0.149 & 0.773 \\
    & Cellulus (size 52) & {\scriptsize unsupervised} & 0.908 & 0.854 & \underline{0.325} & 0.787 \\
    & Routing Pyramids (ours) & {\scriptsize unsupervised} & \underline{0.960} & \underline{0.943} & \textbf{0.649} & \textbf{0.867} \\
    \midrule
    Fluo-HeLa
    & Cellpose-SAM & {\scriptsize supervised} & \textbf{0.915} & \textbf{0.892} & \textbf{0.752} & \textbf{0.843} \\
    & Otsu (size 26) & {\scriptsize non-parametric} & 0.768 & 0.642 & 0.190 & 0.623 \\
    & Cellulus (size 26) & {\scriptsize unsupervised} & 0.875 & 0.824 & 0.297 & 0.756 \\
    & Routing Pyramids (ours) & {\scriptsize unsupervised} & \underline{0.893} & \underline{0.846} & \underline{0.646} & \underline{0.800} \\
    \midrule
    PhC-PSC
    & Cellpose-SAM & {\scriptsize supervised} & \textbf{0.881} & \textbf{0.812} & \textbf{0.094} & \textbf{0.721} \\
    & Otsu (size 19) & {\scriptsize non-parametric} & 0.524 & 0.196 & \underline{0.005} & 0.351 \\
    & Cellulus (size 16) & {\scriptsize unsupervised} & 0.628 & 0.035 & 0.000 & 0.370 \\
    & Routing Pyramids (ours) & {\scriptsize unsupervised} & \underline{0.771} & \underline{0.298} & 0.001 & \underline{0.518} \\
    \bottomrule
  \end{tabular}

\end{table}
\begin{figure}[t]
    \centering
    \begin{subfigure}{\textwidth}
        \includegraphics[width=\textwidth, trim=4.9mm 0mm 3mm 0mm, clip]{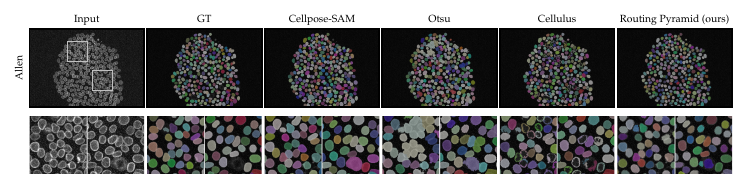}
    \end{subfigure}
    \begin{subfigure}{\textwidth}
        \centering        
        \includegraphics[width=\textwidth, trim=4.9mm 0mm 3mm 0mm, clip]{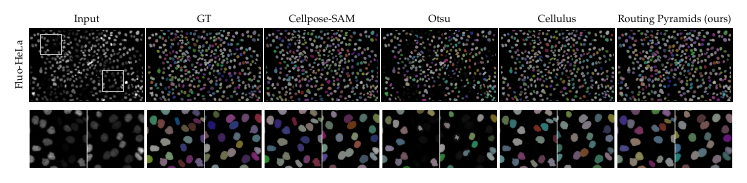}
    \end{subfigure}
    \begin{subfigure}{\textwidth}
        \centering        
        \includegraphics[width=\textwidth, trim=4.9mm 0mm 3mm 0mm, clip]{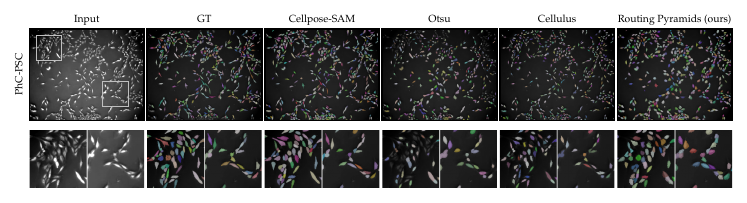}
    \end{subfigure}

    \caption{Instance segmentation on Allen (top), Fluo-HeLa (middle), and PhC-PSC (bottom). 
    From left to right, columns show the
    input image, ground-truth instances (GT), the supervised Cellpose-SAM reference, 
    the unsupervised Otsu and Cellulus baselines, and our method. Instances are
    indicated by distinct colors. The lower row per dataset shows two inset regions marked by white boxes in the input image.
    \label{fig:combined}}
\end{figure} \paragraph{Datasets.}
    We evaluate on three cell culture datasets
    (\cref{tab:segmentation-datasets}):
    the Allen Institute nuclear morphology dataset~\cite{dixonColony2024},
    and the Fluo-N2DL-HeLa and PhC-C2DL-PSC datasets
    from the Cell Tracking Challenge~\cite{maskaCell2023}.
    They span nuclear and whole-cell targets,
    fluorescence and phase-contrast imaging,
    and vary in morphology and density.
    For the Allen dataset, we use the three \emph{analysis} timelapses from
    the \emph{FOV-nuclei} timelapse dataset.
    From these we use the \emph{large} and \emph{small} timelapses for training,
    and the \emph{medium} timelapse for testing.
    For the CTC datasets, we use the two \emph{test} videos (without annotations)
    for training, and the \emph{train} videos (with annotations) for testing.
    As the CTC datasets only have sparse manually annotated segmentation masks,
    we use their \emph{Silver Truth} annotations~\cite{maskaCell2023} as the target.

\paragraph{Training.}
    We train our method from scratch using only single channel images.
    All datasets use the same model architecture with latent dimension $d=64$.
    Models are trained on $256\times256$ crops for 200 epochs with batch size 64.
    We bilinearly upsample the PhC-PSC images by a scale factor of 2.
    The Allen images are maximum-intensity projections of 3D acquisitions,
    and are two-fold downsampled with average pooling.
    We use $\lambda_{\mathrm{sparsity}}=0.5$ for Allen and PhC-PSC,
    and $\lambda_{\mathrm{sparsity}}=0.2$ for Fluo-HeLa.

\paragraph{Baselines.}
    We compare with Cellpose-SAM~\cite{pachitariuCellposeSAM2025}
    (with frozen weights) as a strong supervised reference,
    Cellulus as an unsupervised baseline,
    and Otsu thresholding followed by connected components
    as a non-parametric baseline.
    Cellulus and Otsu use dataset-specific post-processing parameters
    (\eg object size, see \cref{tab:seg_results}) selected on the test set,
    which gives them an advantage in the comparison.
    Our method uses the same post-processing across all datasets
    (\cref{sec:method-inference}).

\paragraph{Metrics.}
    Predicted and annotated instances are matched by intersection over union (IoU).
    We report object-level F1 scores at IoU thresholds $0.5$, $0.7$, and $0.9$,
    and panoptic quality (PQ).

\paragraph{Results.}
    \cref{tab:seg_results} reports the quantitative comparison and
    \cref{fig:combined} shows representative predictions.
    Among the unsupervised methods, \ourmethod obtains the best score on nearly
    every dataset and metric, improving panoptic quality over the strongest
    unsupervised baseline on all three datasets
    (PQ $0.867$, $0.800$, $0.518$ \vs $0.787$, $0.756$, $0.370$ for Cellulus).
    \ourmethod is second only to the supervised
    Cellpose-SAM reference in most settings.
    On the Allen dataset it exceeds this supervised reference in both panoptic
    quality ($0.867$ \vs $0.859$) and F1 at a strict IoU $0.9$ threshold
    ($0.649$ \vs $0.256$), indicating tighter agreement with the ground-truth
    nuclear boundaries.
    The main exception is F1 at IoU $0.9$ on PhC-PSC, where every method scores
    below $0.1$, reflecting the difficulty of pixel-accurate boundaries in
    low-resolution phase-contrast images.

\begin{figure}[t]
  \centering
  \includegraphics[width=\textwidth]{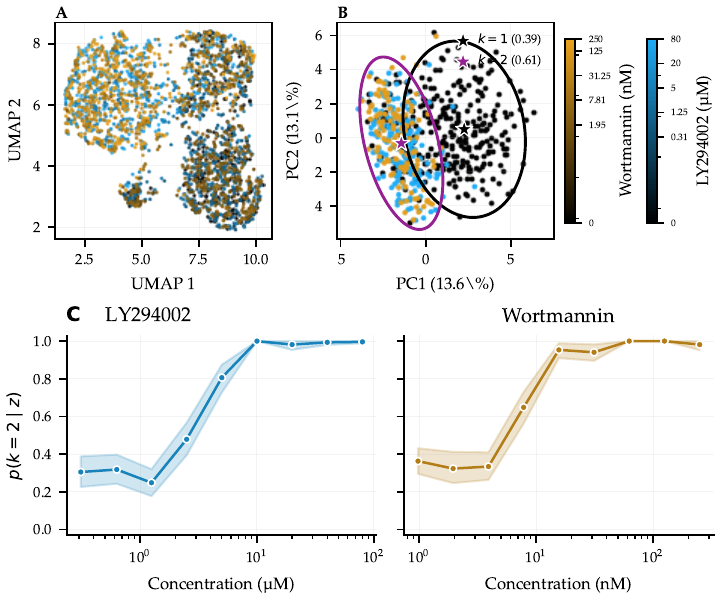}
  \caption{
    Object latents encode perturbation-induced phenotypes.
    a) UMAP of the object latent codes on the series dilution images.
    b) PCA and GMM fit of the training split applied to the control object latents.
    Stars denote the means,
    and elliptical contours show the 95\% confidence intervals (CI)
    for the GMM.
    Legend shows the ratios for each component.
    c) \emph{Positive} (GMM k=2) probability of objects in the series dilution images.
    Lines indicate mean over objects, and the bands indicate 95\% CI.
  }
  \label{fig:u2os_latents}
\end{figure} %
\begin{figure}[t]
  \centering
  \includegraphics[width=\textwidth]{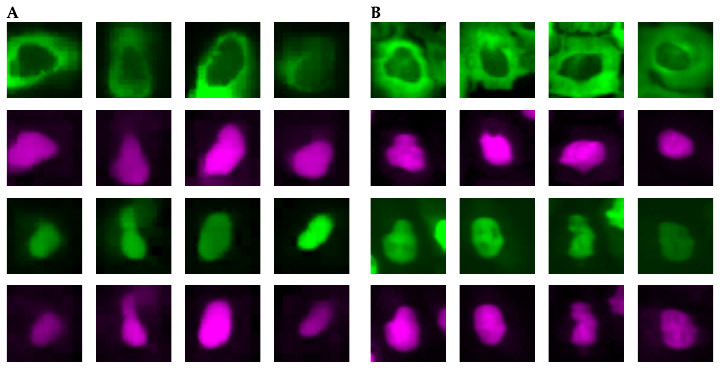}
  \caption{
    Generated (a) and retrieved (b) $32\times32$ crops from the series dilution images.
    Top two rows show GFP (green) and DNA (magenta) channels
    corresponding to the "negative control" GMM component,
    while the bottom two rows correspond to the "positive control" component.
    a) Generated crops by sampling from the two Gaussian distributions.
    b) Retrieved crops by nearest neighbor search with component means
    in the observed object latent set.
  }
  \label{fig:u2os_images}
\end{figure}
 \subsection{Generative Object Representation Learning}
\label{sec:latent-probing}
    We test whether the learned object latent codes capture single-cell phenotypic
    variation using two-channel fluorescence images of U2OS cells treated with a
    dilution series of LY294002 or wortmannin from BBBC013~\cite{ljosaAnnotated2012}.
    Plate columns 2--11 form the series-dilution training split,
    while columns 1 and 12 contain positive and negative controls held out for validation.
    After unsupervised training,
    we extract one embedding per predicted cell (\cref{sec:method-latents}).

    To qualitatively assess latent space clustering by phenotype,
    principal component analysis (PCA) and UMAP~\cite{mcinnesUMAP2018}
    are fit to the series-dilution (training) embeddings,
    and overlaid with perturbation strength (dose) for series dilution (\cref{fig:u2os_latents}a)
    and control images (\cref{fig:u2os_latents}b).
    For both splits, the object latents form distinct clusters by dose.

    To quantitatively model the binary phenotype,
    a two-component Gaussian mixture model (GMM) is
    fit to the object latents from the training split,
    where it shows the dose-response of cells under perturbations (\cref{fig:u2os_latents}c).
    Identities are assigned to the components by comparing their means to the held-out controls
    (\cref{fig:u2os_latents}b).
    Thresholding at 0.5 on the average GMM responsibilities recovers
    the positive/negative classes with 100\% accuracy for the 16 control replicates.

    The GMM can be further used as the approximate posterior for instance generation
    and the clusters for instance retrieval (\cref{fig:u2os_images}),
    by either sampling from the Gaussian components
    or retrieving nearest neighbors (by Euclidean distance) to the component means.
    The generated samples show that the generative model is
    consistent with the phenotypic variation induced by the perturbations,
    while the retrieved samples illustrate the distinct morphologies
    corresponding to the learned latent clusters. %
\section{Conclusion}
\label{sec:conclusion}

We present \ourmethod, an unsupervised object-centric model that learns cell instances
and morphological representations from unlabeled 2D microscopy images.
Its routing-pyramid decoder reconstructs each image
while tracing pixels to spatially sparse latent sources,
so instance masks and object embeddings arise from the same decomposition.

Across fluorescence and phase-contrast datasets,
our method outperforms the evaluated unsupervised baselines
at an IoU threshold of $0.5$, while requiring minimal post-processing.
In a two-channel fluorescence drug perturbation assay,
mixture responsibilities computed from the learned object representations
are consistent with treatment dose,
while generated and retrieved cells recover the corresponding phenotypes.

We currently train a separate model for each dataset,
leaving generalization across diverse samples a question for future work.
Our method also assumes compact objects of similar appearance
against a smoother background,
and application to more complex scenes such as tissue imaging remains unexplored.
Despite these limitations, we show that
recurring visual structure within an experiment can support learning
cell instances and their morphology without manual annotations.
 
\section*{Acknowledgements}
We thank Albert Dominguez Mantes,
École Polytechnique Fédérale de Lausanne (EPFL),
for critically reading the manuscript.

\bibliographystyle{splncs04}

\end{document}